\documentclass{article}
\usepackage{iclr2019_conference,times}

\usepackage[utf8x]{inputenc} 
\usepackage[T1]{fontenc}    
\usepackage{hyperref}       
\usepackage{url}            
\usepackage{booktabs}       
\usepackage{amsfonts}       
\usepackage{nicefrac}       
\usepackage{microtype}      
\usepackage{multirow}
\usepackage{sidecap}
\usepackage{hyperref}       
\usepackage{url}            
\usepackage{booktabs}       
\usepackage{amsfonts}       
\usepackage{nicefrac}       
\usepackage{microtype}      

\usepackage{subfigure}
\usepackage{graphicx}
\usepackage{wrapfig}
\usepackage{listings}
\usepackage{color}
\usepackage{xcolor}
\usepackage{amssymb}
\usepackage{hyperref}
\usepackage{nicefrac}
\usepackage{caption}
\usepackage{bibunits}
\usepackage{amsmath, amssymb}
\usepackage{float}
\usepackage{algorithm}
\usepackage[noend]{algpseudocode}
\usepackage[normalem]{ulem}

\makeatletter
\renewcommand{\ALG@beginalgorithmic}{\small}
\makeatother

\usepackage{graphicx}
\usepackage{nicefrac}
\usepackage{array}
\usepackage{comment}

\usepackage{amsmath, amssymb}
\usepackage{float}
\usepackage{enumitem}

\usepackage[font={footnotesize,it},labelfont={footnotesize,it}]{caption}

\iclrfinalcopy

\title{Neural heuristics for SAT solving}

\author{
  Sebastian Jaszczur\\
  University of Warsaw\\
  \And
  Michał Łuszczyk\\
  University of Warsaw\\
  \And
  Henryk Michalewski\\
  University of Warsaw, deepsense.ai\\
}

\begin{document}

\maketitle

\begin{abstract}
We use neural graph networks with a message-passing architecture and an attention mechanism to enhance the branching heuristic in two SAT-solving algorithms. We report improvements of learned neural heuristics compared with two standard human-designed heuristics.
\end{abstract}

\section{Introduction}
\label{sec:introduction}

\begin{wrapfigure}{r}{0.5\textwidth}
\vspace{-20pt}
\begin{minipage}{.5\textwidth}
\centering
\begin{algorithm}[H]
\begin{algorithmic}[1]
\Function{dpll}{$\Phi$}
\State $\Phi \gets \texttt{simplify}(\Phi)$
\If {$\Phi \text{ is trivially satisfiable}$} \Return True
\EndIf
\If {$\Phi \text{ is trivially unsatisfiable}$} \Return False
\EndIf
\State $literal \gets \texttt{choose-literal}(\Phi)$
\If {DPLL($\Phi \land literal$)} \Return True \EndIf
\If {DPLL($\Phi \land \neg literal$)} \Return True \EndIf
\State\Return False
\EndFunction
\end{algorithmic}
\end{algorithm}
\end{minipage}
\renewcommand{\figurename}{Algorithm}
\caption{High level overview of DPLL. In this work, we embed neural network as \texttt{choose-literal}. In DPLL, \texttt{simplify} contains unit propagation and clause elimination. Trivially satisfiable and trivially unsatisfiable for CNF means respectively an empty formula, and a formula containing an empty clause. }
\label{fig:main_loop}
\label{backtracking}
\vspace{-15pt}
\end{wrapfigure}

The Boolean satisfiability problem (SAT) is the problem of determining the existence of a solution for a given propositional logic formula. It is a NP-complete problem, meaning that any NP problem can be reduced to SAT problem in polynomial time \cite{SATisNP}.

We explore the possibility of using neural networks in SAT solving as branching heuristics in search algorithms\footnote{See \url{https://bit.ly/neurheur} for a TPU-bound implementation of all algorithms in this paper.}. 
We focus on two  SAT-solving algorithms: 
DPLL (Algorithm \ref{backtracking}) and more advanced 
CDCL. Both of those are complete backtracking-based search algorithms. Both depend on the branching heuristic \texttt{choose-literal}, which chooses branching variable and its Boolean value. 
The expected running time is heavily dependent on the quality of this heuristic \cite{MarquesSilva}. In this work we use neural networks as \texttt{choose-literal} heuristic and compare its performance with DLIS and Jeroslow-Wang One-Sided (JW-OS) heuristics, which are presented in \cite{MarquesSilva, CHAFF} as one of the best strategies in most circumstances. We compare the performance in terms of number of branching decisions and show the possibility of enhancing the performance of SAT solvers with the help of learned heuristics.

\section{Related work}
\label{sec:related_work}
\cite{neurosat} proposed the {\bf NeuroSAT} message-passing network architecture for SAT-solving that generates the assignment of variables directly in the graph. This is an important inspiration for our work, although in contrast to \cite{neurosat}  we use a message-passing network for guidance of a backtracking-based algorithm instead. A similar graph representation, but more general in order to accommodate for higher-order logic is used in \textbf {FormulaNet} presented in \cite{formulanet}. To the best of our knowledge the FormulaNet architecture was never used for neural guidance. In \cite{neurosat, formulanet} formulas are represented as graphs and a general approach to neural networks and graphs, including the attention mechanism, can be found in 
\cite{InductiveBias, GAT}. 
The {\bf PossibleWorldNet} architecture described in \cite{entailment} is based on the {\bf TreeNN} architecture, with an additional idea of checking multiple possible worlds. We consider the exploration of possible worlds as an alternative to structured backtracking-based search algorithms like DPLL and CDCL. 
It is worth noting that PossibleWorldNet could be modified to use a message-passing architecture while keeping the exploration of possible worlds. Another application of TreeNN for proof synthesis in propositional logic was proposed in \cite{proofsynthesis}. 
{\bf EqNet} \cite{eqnet} solves a more general problem of determining equivalence of Boolean (alternatively: arithmetic) expressions (satisfiability can be seen as equivalence to any unsatisfiable formula e.g. \(a \land \neg a\)). However, the formulas solved by EqNet have up to 10 variables and 13 symbols, while we tackle formulas beyond one hundred variables and thousands of symbols.
{\bf Learned Restart Policy} \cite{restart-policy} presents a different approach to improve a SAT solver with machine learning, where the network decides at each step whether the algorithm should be restarted to follow another random path in the search tree.
\cite{learnedclause} uses reinforcement learning to train clause deletion heuristics in DPLL based solvers.
\\

\vspace{-10pt}
\section{Architecture}
\label{sec:architecture}
\begin{figure}[H]
\begin{minipage}{.45\textwidth}    
  \centering
  \includegraphics[width=2.2in]{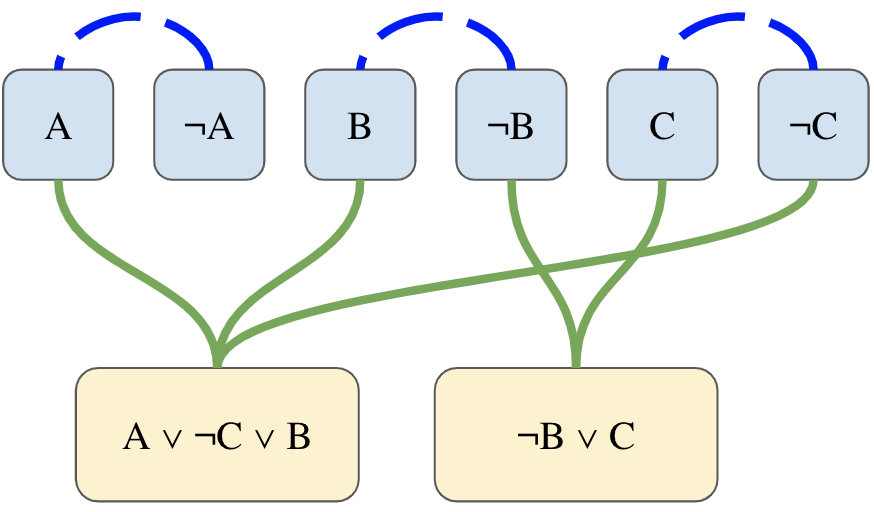}
  \label{fig:main_neuro}
\end{minipage}%
\hfill
\begin{minipage}{.55\textwidth}
  \centering
  \includegraphics[width=3in]{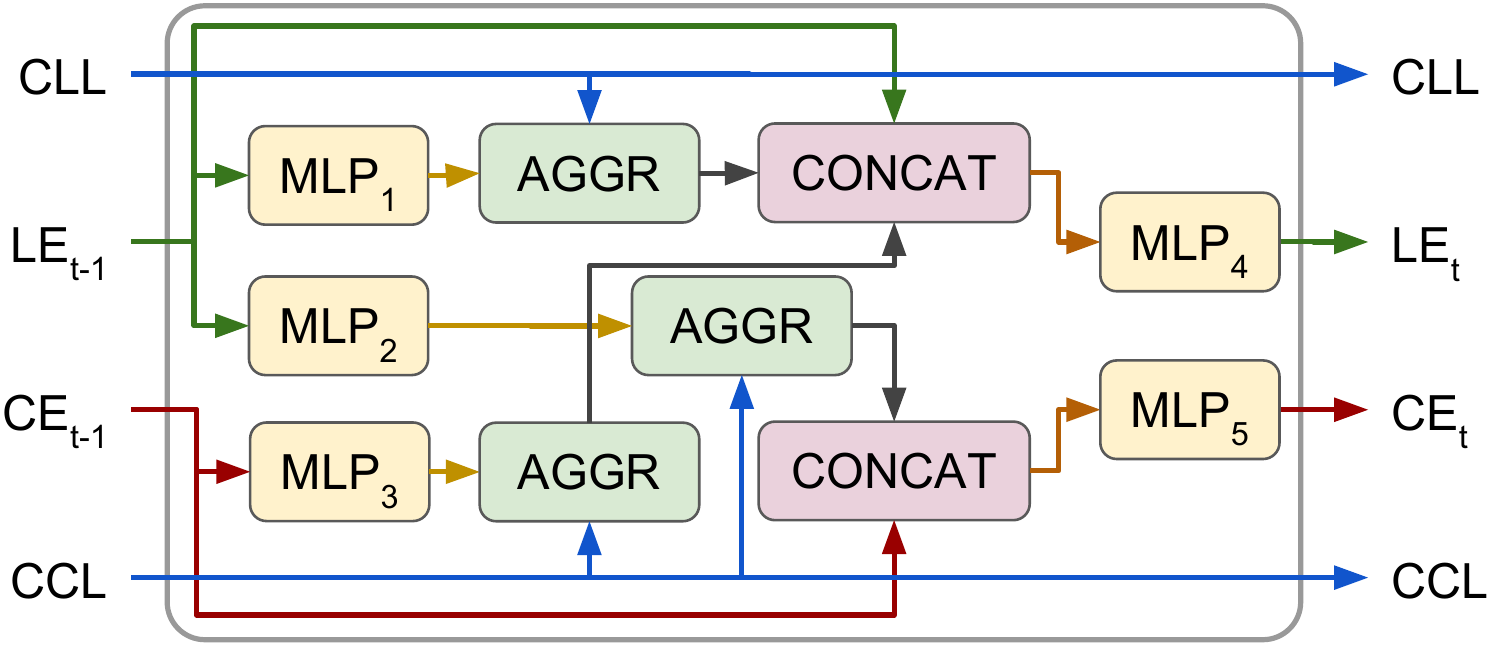}
  \label{fig:rnncell}
\end{minipage}
\caption{Left: A graph representation of formula $(A \lor \lnot C \lor B) \land (\lnot B \lor C)$ used in our work. In the model nodes are unlabeled (labels are included only for the reader's convenience). Different colors mark two distinct types of nodes (clause and literal) and two distinct types of edges (literal-literal and clause-literal). Right: Overview of message-passing architecture. In each iteration we take as the input: connection matrix between clauses and literals (\(CCL\)), connection between literals and their negations (\(CLL\)), literal embeddings from previous iteration (\(LE_{t-1}\)), and clause embeddings from previous iteration (\(CE_{t-1}\)). We use $5$ separate MLPs, which share parameters across iterations. Aggregation method depends on a model, see the description below.}
\label{fig:architecture}
\end{figure}

We use a message-passing graph-based neural network architecture similar to NeuroSAT introduced in \cite{neurosat}. The general idea is to represent a formula as a graph with two node types (literal and clause) and two edge types (literal-literal edges represent the negation relation, and clause-literal edges represent relation between each clause and literals it contains). Example formula represented as a graph is shown in Figure \ref{fig:architecture} Left. Each node has its own state, represented by an embedding vector. Thanks to this representation, we have the following properties: \textbf{1.} Invariance to variable renaming. \textbf{2.} Invariance to negation of all occurrences of a variable. \textbf{3.} Invariance to permutation of literals in a clause. \textbf{4.} Invariance to permutation of clauses in a formula. 

We initialize all embedding vectors with a trainable initial embedding, different for each type of node. Then we run a number of iterations (from $20$ to $40$ in our experiments), visualized in Figure \ref{fig:architecture} Right. Each iteration consists of three stages:
\textbf{Stage 1.} Message: Each node generates a message vector \(V\) (and a vector \(K\) if needed) based on its embedding, to every connected node. \(V\) and \(K\) are generated with a  three-layer MLP with LeakyReLU \cite{LReLU} activation after each hidden layer and linear activation after the last layer.
\textbf{Stage 2.} Aggregate: all messages are delivered according to the connection matrix, then aggregated for each receiver with one of the aggregation functions (described in the next paragraph).
\textbf{Stage 3.} Update: Each node updates its embedding based on its previous embedding and aggregated received messages. New embedding is computed by a three-layer MLP with LeakyReLU activation after each hidden layer and sigmoid activation after the last layer.

We explore two different aggregation methods. The first is the average of received \(V\) vectors.
The second method is a modified attention mechanism. As a message, instead of just a single vector \(V\), we send two vectors, \(V\) and \(K\). Receiving node generates one vector \(Q\) based on its embedding, and the result of aggregation is \(\sum_{i} V_i\ sigmoid(K_i \cdot Q)\). Thanks to this, each message may be selectively rejected or accepted by the receiver, depending on relation between \(K\) and \(Q\). The intuitive difference between this mechanism and the standard attention is as follows: the standard attention as in \cite{GAT} chooses one message to look at, while our mechanism rejects or accepts messages independently and looks at their sum.

Like NeuroSAT, our architecture learns to predict satisfiability of the whole formula (which we name \textit{sat prediction}). However, it also predicts, for each literal separately, the existence of a solution with this literal (which we name \textit{policy prediction}).
To get \textit{policy prediction} we add a logistic regression on top of each literal's embedding in each iteration (with parameters shared across all literals and iterations). To get \textit{sat prediction} we add a linear regression on top of each literal's embedding in each iteration, and then apply a sigmoid on sum of their outputs. We define \textit{sat loss} as cross-entropy loss %
between the \textit{sat prediction} and the ground truth. We define \textit{policy loss} as zero if formula is unsatisfiable and as the average of cross-entropy losses between \textit{policy predictions} and ground truths if formula is satisfiable. To get a loss of the model we sum together both losses for every iteration.

\vspace{-10pt}
\section{Experimental results}
\label{sec:experimental}
\paragraph{Dataset and training details.}
To train and evaluate the models we use a class of SAT problems $SR(n)$ introduced and described in detail in \cite{neurosat}. It is parametrized only by $n$ – the number of variables used in a formula. Both the size and the number of clauses vary. The dataset is balanced in terms of number of satisfiable and unsatisfiable examples. Each of the $SR(n)$ samples has two labels (see Section \ref{sec:architecture}): {\it sat} indicating whether the formula \(\Phi\) is satisfiable and {\it policy} indicating for each literal \(l\) whether \(\Phi \land l\) is satisfiable. We generate each of those numbers by running
MiniSat 2.2 \cite{minisat}. Sample random $SR(30)$ formulas are solved by MiniSAT 2.2 in 0.007 seconds, while $SR(110)$ takes 0.137 second and $SR(150)$ takes 3.406 seconds (for a Xeon E5-2680v3@2,5 GHz computer). We have trained separate models on SR(30), SR(50), SR(70) and SR(100). Table \ref{tab:training} shows the details of the training procedure. Metrics \(sat\) error and \(policy\) error are defined as mean absolute error of \(sat\) or \(policy\) prediction versus labels. The presented models are message-passing neural networks with our modified attention mechanism.

\begin{table}[H]
\vspace{-10pt}
\centering
\resizebox{\textwidth}{!}{%
\begin{tabular}{lrrrrrr}
Problem & Loss & {\it sat} error & {\it policy} error & Batch size &  Train. steps & Train. time \\
\hline
 SR(30) & 28.178±0.672 & 0.084±0.004 & 0.050±0.002 & 128 & 1200K & 20h \\
 SR(50) & 32.024±0.555 & 0.233±0.017 & 0.105±0.006 & 64 & 600K & 12h \\
 SR(70) & 33.010±0.482 & 0.266±0.033 & 0.110±0.007 & 64 & 600K & 22h \\
SR(100) & 34.227±0.127 & 0.319±0.007 & 0.123±0.002 & 32 & 1200K & 28h \\ 
\end{tabular}}
\caption{Each of the models was trained on SAT samples drawn from the distribution marked in the first column. The metrics: loss, \textit{sat} error and \textit{policy} error are evaluated on an independently generated evaluation set. The values indicate mean and standard deviation over 3-5 trained models. Models were trained using single TPU v2.} 
\label{tab:training}
\vspace{-20pt}
\end{table}

\begin{wrapfigure}{r}{0.45\textwidth}
\vspace{-10pt}
\centering
\includegraphics[width=6cm]{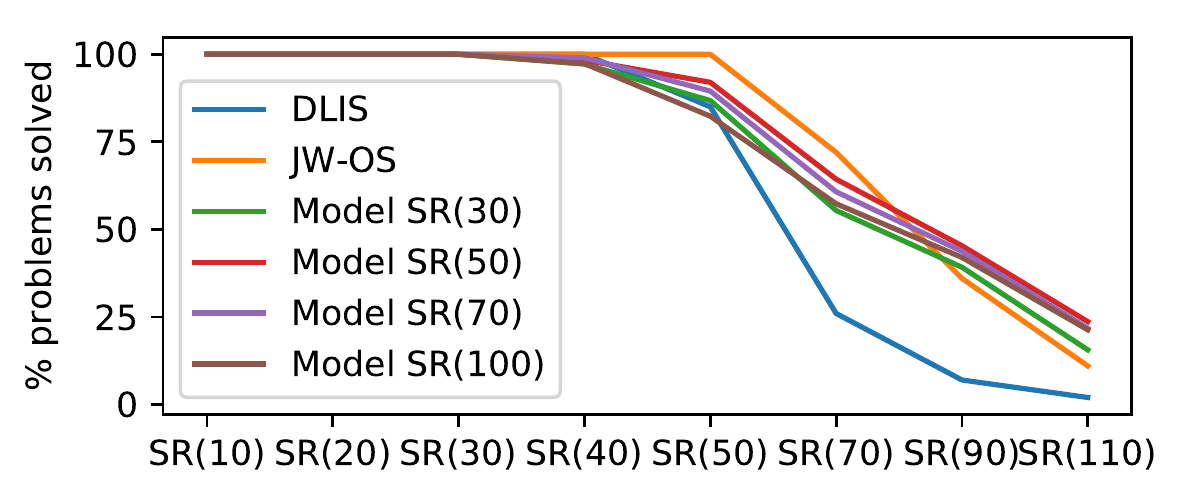}
\caption{Performance of DPLL with different guidance heuristics on specific problem sizes. The \(x\) axis indicates the class of the evaluation set: evaluation is performed on fresh randomly chosen one satisfiable hundred $SR(x)$ formulas. The \(y\) axis indicates the percent of instances (out of 100) solved by DPLL within 1000 steps.}
\label{fig:1e3steps}
\vspace{-5pt}
\end{wrapfigure}

\paragraph{Experiment 1: comparison of all models with DLIS and JW-OS heuristics.}
\label{sec:srn-results}
We evaluated the DPLL algorithm guided by our 4 kinds of models described above and compared to DPLL guided by JW-OS and DLIS. As a performance consideration we decided to  stop  DPLL after 1000 steps (see Experiment 2 below for a comparison without this restriction) and count the number of solved formulas out of 100 in each class.  We present the results in Figure \ref{fig:1e3steps}. For this and subsequent experiments we only consider satisfiable $SR(n)$ samples. JW-OS proved to be the best on average classes of problems: $SR(50)$ and $SR(70)$, whereas neural guidance-based algorithms proved to be the best on large problems: $SR(90)$ and $SR(110)$. 

\begin{figure}[H]
\begin{minipage}{.5\textwidth}
\centering
\includegraphics[width=7cm]{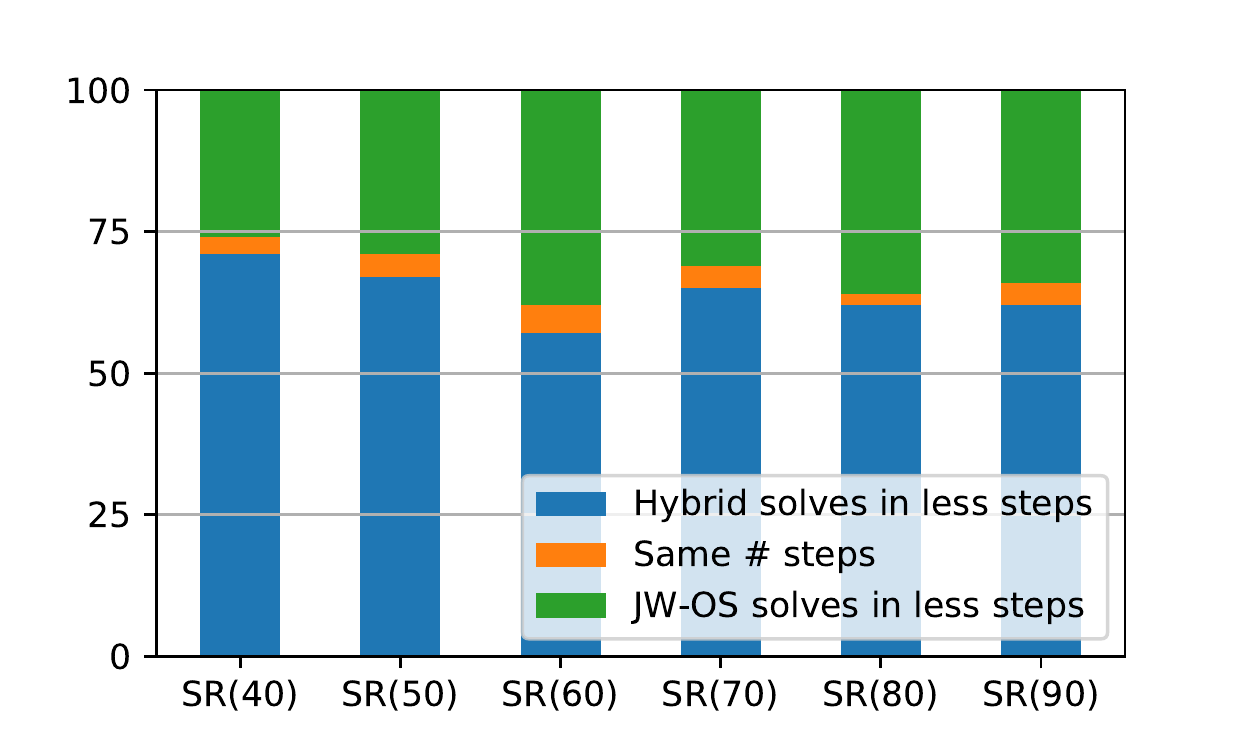}
\end{minipage}
\hfill
\begin{minipage}{.5\textwidth}
\centering
\includegraphics[width=7cm]{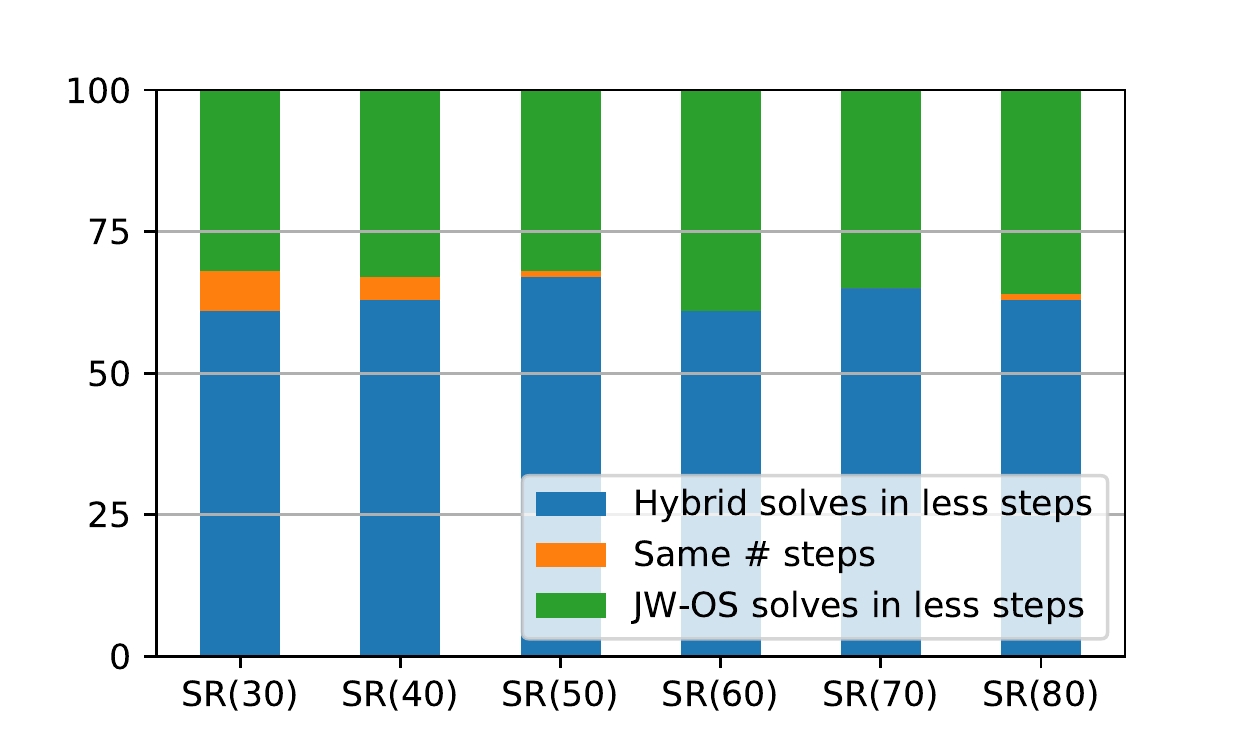}
\end{minipage}
\caption{Left: Comparison of Hybrid (ours) and JW-OS as heuristics in DPLL. We measure the performance of each method according to the number of steps required to find a solution for a given SAT instance. A method wins if it solves a given instance in a smaller number of steps. The blue bar reflects the percentage of formulas where the Hybrid (ours) method won, the green bar means that JW-OS won, and the orange bar means that there was a draw.  Right: the same for CDCL.}
\label{fig:less-steps}
\end{figure}

\paragraph{Experiment 2: detailed comparison with the JW-OS heuristic.}
We have selected the SR(50) model for a detailed comparison of the learned heuristics versus JW-OS and for the sake of this comparison designed 
hybrid guidance algorithm that uses a model trained on $SR(50)$ (a fixed one of the three similar replicas) and switches to JW-OS when the network predicts \(sat\) probability below a threshold of $0.3$\footnote{We leave further parameter and model searches as a topic which should be considered in the full version of this paper.}.
We then compared the new hybrid guidance with the heuristic JW-OS without the 1000 step restriction. JW-OS was selected on the basis of Experiment 1.
The experiment shows that the hybrid approach is faster in terms of  number of steps in a significant majority of cases, both when used with DPLL (Figure \ref{fig:less-steps} Left) and with CDCL (Figure \ref{fig:less-steps} Right).

\begin{wrapfigure}{r}{0.45\textwidth}
    \vspace{-30pt}
    \centering
    \includegraphics[width=7cm]{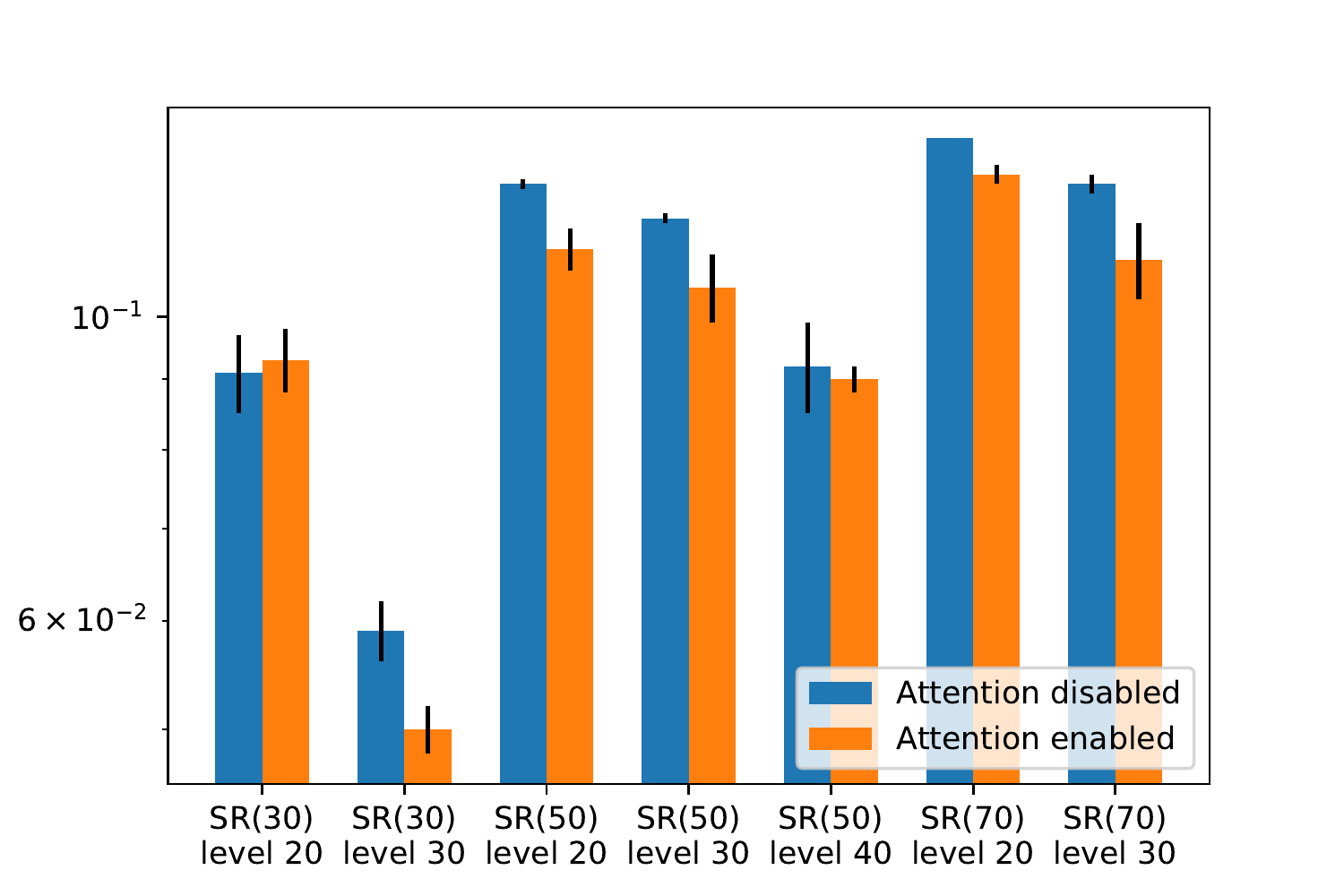}
    \caption{Comparison of \textit{policy} error with and without attention. The presented values are mean and standard deviation over 3-5 trained models calculated on the evaluation set.}
    \label{tab:attention}
    \vspace{-20pt}
\end{wrapfigure}

\paragraph{Experiment 3: an ablation for the attention mechanism.}
From experiments presented in Figure \ref{tab:attention} follows that in most cases attention improved evaluation metrics by a significant margin. Only in the case of $SR(30)$, level 20 attention degraded the model performance. For $SR(50)$, level 40 the metrics with and without attention stayed within the standard deviation of each other.

\paragraph{Reproducibility.}
\label{reproducibility}
\label{sec:reproducibility}
For each set of hyperparameters (e.g. $SR(30)$, level 40), we trained five models.   
We considered a model not correctly trained if adding it to the set of models raised standard deviation of the losses above 1, see Table \ref{tab:training}. We excluded such models (up to $2$ models out of $5$ for a set of hyperparameters) from  further comparisons and left the question of stability of training as a topic of further investigations. 
The code 
including hyperparameters is published at \url{https://bit.ly/neurheur}.  
Our code is based on TensorFlow \cite{tensorflow2015-whitepaper}. It uses a CDCL implementation by \cite{cdcl-implementation}. We access MiniSat through PySAT interface \cite{python-sat}.

\section{Conclusions and future work}
\label{sec:conclusions}
\label{sec:future}
In this work we have shown three experiments confirming that SAT-solving can be augmented by neural networks. The message-passing architecture augmented by attention performs competitively comparing with standard heuristics when evaluated on relatively large propositional problems, including problems with more than a hundred variables (see Section \ref{sec:srn-results}). From the ablation presented in Experiment 3 follows that the message-passing architecture that uses the attention mechanism overall performs better then the same architecture without attention and we attribute it to a selective acceptance of incoming messages made possible by the attention mechanism.
We believe that using an appropriately large computing infrastructure the learning process can be extended to more complex examples and that in the near future parallelization combined with a variant of the message-passing architecture can be used to train models which will tackle larger SR problems, and possibly SAT problem classes currently beyond the reach of SAT-solvers.
As a future step we consider extending our improved heuristics so that a  neural network would be able to control other aspects of the SAT solver behavior, like restarting and backtracking.
Eventually, other prediction targets, including expected number of steps, may be beneficial. Once we exhaust the pool of available supervised data it would be interesting to apply reinforcement learning methods, including methods recently presented in  \cite{rlcop}. 
In this work we focus on the number of steps of the algorithm rather than execution time. Moving the main loop of DPLL or CDCL to a tensor computation graph would be a step towards making the algorithms more competitive in terms of the execution time.

\section{Acknowledgements}
This was work was supported by (1) the Polish National Science Center grant UMO-2018/29/B/ST6/02959 (2) the TensorFlow Research Cloud which granted 50 TPUs  (3) the  Academic Computer Center Cyfronet at the AGH University of Science and Technology in Kraków, Poland.

\bibliography{deepsat}

\newpage
\appendix

\end{document}